\documentclass[runningheads]{llncs}
\usepackage[T1]{fontenc}
\usepackage{graphicx,verbatim}

\usepackage{comment}
\usepackage{amsmath}
\usepackage{amssymb}
\usepackage{diagbox}
\usepackage{multirow}
\usepackage{threeparttable}
\usepackage{xcolor}
\usepackage{booktabs}
\usepackage{placeins}
\usepackage{bm}
\usepackage{array}
\usepackage{tabularx}
\usepackage[export]{adjustbox}
\begin{document}
\title{Meta-D: Metadata-Aware Architectures for Brain Tumor Analysis and Missing-Modality Segmentation}
\titlerunning{Meta-D: Metadata-Aware Architectures for Analysis and Segmentation}
%

\author{SangHyuk Kim\inst{1} \and
Daniel Haehn\inst{1} \and Sumientra Rampersad\inst{1}}
\authorrunning{S. Kim et al.}
%
\institute{University of Massachusetts Boston\\
\email{sanghyuk.kim001, daniel.haehn, sumientra.rampersad@umb.edu}}
\maketitle              

\begin{abstract}




We present Meta-D, an architecture that explicitly leverages categorical scanner metadata such as MRI sequence and plane orientation to guide feature extraction for brain tumor analysis. We aim to improve the performance of medical image deep learning pipelines by integrating explicit metadata to stabilize feature representations. We first evaluate this in 2D tumor detection, where injecting sequence (e.g., T1, T2) and plane (e.g., axial) metadata dynamically modulates convolutional features, yielding an absolute increase of up to 2.62\% in F1-score over image-only baselines. Because metadata grounds feature extraction when data are available, we hypothesize it can serve as a robust anchor when data are missing. We apply this to 3D missing-modality tumor segmentation. Our Transformer Maximizer utilizes metadata-based cross-attention to isolate and route available modalities, ensuring the network focuses on valid slices. This targeted attention improves brain tumor segmentation Dice scores by up to 5.12\% under extreme modality scarcity while reducing model parameters by 24.1\%.

\keywords{metadata  \and segmentation \and transformer}

\end{abstract}
\section{Introduction}

Multi-parametric MRI captures brain tissue properties using sequences such as T1, which highlights fat and detail, and FLAIR, which shows lesions by suppressing fluid signals. Medical deep learning models use these 3D pixel arrays to automatically segment and classify brain tumors. Each scan includes categorical metadata about its imaging sequence (scan type) and spatial plane (orientation: axial, sagittal, or coronal). Standard neural networks often ignore this information and rely only on image textures to infer scanner details. This implicit method can cause contrast ambiguity when sequences have overlapping visual intensities, making tissue differentiation difficult.

Existing methods address brain tumor analysis and missing-modality challenges with different architectural designs. Some frameworks add metadata through late fusion at the final classification layers~\cite{fusion1review,fusion2review}. Others employ advanced recalibration or localized attention. For example, frequency-domain information recalibrates features~\cite{sffr}. Other architectures use shifted windows or specialized attention to separate features~\cite{swintransformer,huang2022missformer}. Some techniques use dynamic spatial routing to reduce complexity based on image content~\cite{routingtransformer}. To address missing sequences, architectures often apply spatial zero-padding with multimodal self-attention~\cite{mmformer,transbts}. Although these methods successfully fuse representations, they rely entirely on visual image content to calculate attention pathways. They do not explicitly use categorical metadata to govern routing. As a result, self-attention still processes spatial domains with zero-padded noise when structural sequences are missing.

In this paper, we present Meta-D, a neural architecture that uses categorical metadata to guide feature extraction. We propose a two-part framework to evaluate this approach. In 2D tumor classification, Meta-D applies feature-wise linear modulation (FiLM)~\cite{film}. FiLM adjusts neural network features using conditioning information. Here, FiLM scales intermediate features using only sequence and plane metadata. This explicit use of metadata resolves image contrast ambiguity and improves classification performance.

\begin{figure}[h!]
\centering

\includegraphics[width=1\textwidth]{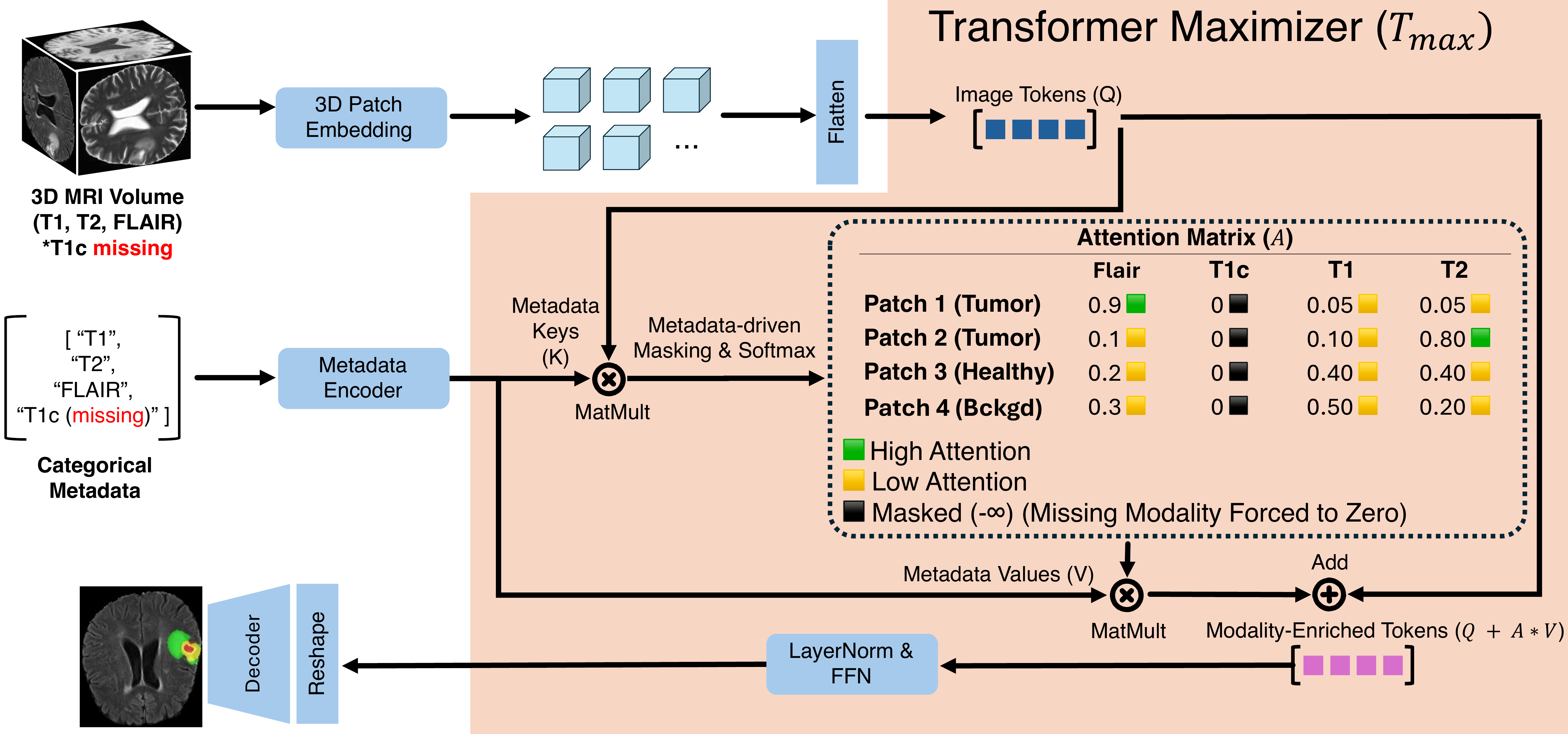}

\caption{An overview of the Meta-D ($T_{max}$) architecture. By querying a fixed 4-modality metadata dictionary instead of $N$ spatial patches, the module handles missing sequences via deterministic masking and reduces attention complexity from $O(N^2)$ to $O(N)$.}
\label{fig:overview}
\end{figure}

Because categorical metadata anchors feature representations with complete data, we test its stability when imaging sequences are missing. We extend our framework to 3D missing-modality tumor segmentation. Our 3D module, the Transformer Maximizer (Meta-D $T_{max}$), uses metadata-driven cross-attention (see Fig.~\ref{fig:overview}). The architecture queries a fixed metadata dictionary to handle missing data without spatial inference over empty regions. In the absence of data, Meta-D delivers efficient models. It reduces the model parameter footprint by 24.1\% and elevates 3D tumor segmentation scores.

\section{Methodology} 

\subsection{Metadata-Guided 2D Brain Tumor Classification}
To resolve contrast ambiguity between MRI sequences (e.g., misinterpreting bright fluids in T2 as contrast agents in T1c, or confusing bright fat in T1 with edema in FLAIR) and geometric variance across imaging planes (e.g., confusing the long, vertical shape of structures in sagittal views with their wider cross-sections in axial views), the network is explicitly conditioned prior to feature extraction. We achieve this using FiLM~\cite{film}. A dedicated multi-layer perceptron (MLP) maps discrete metadata strings, representing the sequence type and anatomical plane, into continuous scaling ($\gamma$) and shifting ($\beta$) vectors. These vectors dynamically modulate intermediate convolutional feature map $x$:$$FiLM(x_c) = \gamma_c x_c + \beta_c$$
where $c$ denotes the specific channel index. This explicitly forces the 2D Meta-D encoder to recalibrate its feature extraction based on both the physical contrast of the scanner modality and the spatial geometry of the anatomical plane. The modulated output is globally pooled and passed to a classification head for binary tumor classification.

\subsection{3D Brain Tumor Segmentation} We hypothesize that if categorical metadata reliably grounds the feature space, it should successfully guide the network even when certain image sequences are absent. Building upon the performance improvements observed in 2D classification, we investigate the scalability of explicit metadata injection to 3D volumetric tasks. The missing-modality segmentation scenario provides an optimal environment to test this capability. To leverage this without relying on computationally heavy 3D zero-padding, we introduce a deterministic, metadata-driven routing mechanism: the Transformer Maximizer ($T_{max}$) block.

\subsubsection{Image and Metadata Tokenization}

The surviving 3D MRI volumes are partitioned into patches, flattened, and projected with positional embeddings to form the spatial query matrix $Q \in \mathbb{R}^{N \times D}$, where $N$ is the number of spatial image patches and $D$ is the embedding dimension. Conversely, the key matrix $K \in \mathbb{R}^{M \times D}$ and value matrix $V \in \mathbb{R}^{M \times D}$ are generated exclusively from a predefined, fixed-size metadata dictionary (T1, T1c, T2, FLAIR) via a distinct metadata encoder, where $M=4$ represents the maximum number of sequence modalities.

\subsubsection{The Transformer Maximizer ($T_{max}$) Block}

Raw attention logits between the spatial image tokens ($Q$) and the categorical metadata tokens ($K$) are computed as $S=\frac{QK^T}{\sqrt{D}}$. To handle absent sequences deterministically, $T_{max}$ introduces an explicit masking matrix $M\in\mathbb{R}^{N\times M}$. If modality $j$ is missing, the corresponding column in $M$ is populated with $-\infty$; if available, it is populated with 0. This mask is added directly to the raw logits:$$A=\text{Softmax}(S+M)=\frac{e^{(S_i+M_i)}}{\sum e^{(S_j+M_j)}}$$Because the softmax function exponentiates the inputs ($e^{-\infty}=0$), the attention probabilities for any missing modality are mathematically forced to exactly zero. The attended metadata values are computed by multiplying this masked attention matrix by the metadata values ($AV$). These values are then added back to the original spatial image tokens via a residual connection to form the final modality-enriched spatial tokens:$$\text{Modality-Enriched Tokens}=Q+A*V$$ This functions as a dual-action mechanism: (1) It provides semantic routing by actively matching spatial patches to the most useful available modality, and (2) it performs deterministic masking by physically severing the mathematical linkages to absent sequences. Consequently, values from missing modalities cannot corrupt the spatial feature map. Furthermore, by extracting $K$ and $V$ strictly from the discrete metadata space rather than the spatial domain, attention complexity is reduced from a quadratic $O(N^2)$ to a linear $O(N\cdot M)$.

Following the $T_{max}$ block, the enriched 1D sequence passes through standard layer normalization and a feed-forward network. The sequence is reshaped into a 3D spatial grid and processed by a 3D convolutional neural network (CNN) decoder to generate the final segmentation masks, as depicted in Fig.~\ref{fig:overview}.

\section{Experiments}

\subsection{Datasets and Preprocessing}
\textbf{2D Tumor Classification} We utilized the BraTS 2020 dataset~\cite{bratsref1,bratsref2,bratsref3} comprising 369 subjects for training, internal validation, and testing. The BRISC dataset~\cite{brisc} was employed as an external test set. Data was split by subject IDs to guarantee zero subject overlap across all subsets and datasets. To eliminate human annotation errors, plane orientation metadata was programmatically extracted from RAS coordinates in NIfTI headers, and sequence metadata was obtained directly from the official dataset annotations. Binary labels of \textit{Glioma} versus \textit{No Tumor} were assigned based strictly on the presence of segmented tumor voxels in the ground truth masks, requiring a minimum of 30 pixels to align with the lower limit of human visual recognition. We exclusively utilized T1 and T2 sequences because their highly distinct structural and fluid-attenuated contrast profiles provide an optimal baseline to isolate the impact of metadata injection. 

During slice extraction, a voxel coverage threshold of 0.2 was applied to ensure at least 20 percent of valid tissue voxels were included. An intensity threshold of 0.1 was applied to avoid excessively dark slices. To mitigate magnetic field inhomogeneities stemming from different scanner types, we applied N4 bias field correction~\cite{biascorrection} and evaluated both uncorrected and bias-corrected variations. To minimize anatomical redundancy and capture distinct structural differences per subject, we extracted every 10th slice. Following all filtering steps, the \textit{No Tumor} class yielded more valid slices than the \textit{Glioma} class. To eliminate class imbalance effects and focus purely on metadata efficacy, we randomly sampled 5500 images per class to establish a perfectly balanced dataset at the maximum available capacity. Because the BraTS 2020 training data is skull-stripped by default, we evaluated the models on the BRISC~\cite{brisc} test cohort under both untouched and skull-stripped conditions via FSL BET~\cite{fslbet} to explicitly verify structural robustness.

\textbf{3D Tumor Segmentation} To ensure a fair architectural comparison against the baseline~\cite{mmformer}, we adopted their exact data preprocessing and partitioning setup utilizing the BraTS 2018 dataset~\cite{bratsref1,bratsref2,bratsref3}. All 3D volumes comprising T1, T1c, T2, and FLAIR sequences were co-registered, interpolated to 1 millimeter isotropic resolution, skull-stripped, and independently z-score normalized using only non-zero brain voxels. Zero-value spatial background regions were tightly cropped to eliminate redundant computational overhead. Data partitioning precisely matched the predefined baseline splits. To accommodate the memory constraints of an NVIDIA RTX 2080 Ti GPU, both architectures processed $64 \times 64 \times 64$ input crops with a batch size of 1 and an internal patch size of 4.

\subsection{Implementation Details}
\textbf{2D Meta-D} A ResNet-18~\cite{resnet} backbone was modulated via Progressive FiLM~\cite{film}. This progressive strategy injects explicit metadata solely into deeper semantic layers, preserving the integrity of early primitive feature extractors. Discrete sequence and plane metadata were embedded into 16-dimensional vectors, concatenated into a 32-dimensional context vector, and processed by a 64-unit MLP to predict affine scaling $\gamma$ and shifting $\beta$ parameters. Modulation was injected at Layer 3 with 256 channels and Layer 4 with 512 channels, utilizing a residual formulation of $x + \gamma x + \beta$. We applied standard spatial augmentations including resize to 256, crop to 224, horizontal flip, rotation up to 15 degrees, and color jitter of 0.2. Models were optimized using the Adam optimizer~\cite{adamoptimizer} with a learning rate of 1e-4, a weight decay of 1e-4, and a batch size of 32. We monitored the validation loss and accuracy uniformly across all configurations and set the evaluation limit to 50 epochs.

\textbf{3D Meta-D ($T_{max}) $} Training proceeded for 1000 epochs using the Adam optimizer~\cite{adamoptimizer} with a learning rate of 2e-4, a weight decay of 1e-4, and gradient clipping at 1.0. The objective function combined cross-entropy and Dice loss. Auxiliary deep supervision branches, which are used to guide the gradient flow of intermediate layers during training, were decayed by a factor of 0.4 in later epochs.

\subsection{Results and Discussion}

\newcolumntype{Y}{>{\centering\arraybackslash}X}

\begin{table}[htbp]
\centering
\caption{2D Tumor Detection F1-Score Performance. Models were trained on balanced 5500 slice subsets. Values are presented as mean $\pm$ std. on three-fold cross validation. Best results per scenario are bolded, showing that Meta-D with both sequence and plane metadata consistently outperforms all other models.}
\label{tab:2d_film}
\scalebox{0.88}{ 
    \begin{tabularx}{1.13\textwidth}{>{\centering\arraybackslash}p{2.2cm}  c  c  Y Y Y}
    \toprule
    \multirow{3}{*}{\textbf{Dataset}} & \multirow{2}{*}{\textbf{bias-}} & \multirow{2}{*}{\textbf{2D Image}} & \multicolumn{3}{c}{\textbf{Meta-D (Ours)}} \\
    \cmidrule(lr){4-6}
    & \textbf{corr.} & \multirow{1}{*}{\textbf{only}} & \multirow{2}{*}{\textbf{Img + Seq}} & \multirow{2}{*}{\textbf{Img + Plane}} & \textbf{Img + Seq + Plane} \\
    \midrule
    
    \multirow{2}{*}{BraTS 2020} & \checkmark & $0.9037 \pm 0.0033$ & $0.9129 \pm 0.0015$ & $0.9108 \pm 0.0255$ & \textbf{0.9138} $\pm 0.0089$ \\ 
     &  & $0.9055 \pm 0.0209$ & $0.9030 \pm 0.0130$ & $0.9070 \pm 0.0028$ & \textbf{0.9099} $\pm 0.0134$ \\ \addlinespace
    
    \multirow{2}{*}{BRISC} & \checkmark & $0.6890 \pm 0.0226$ & $0.6860 \pm 0.0622$ & $0.6955 \pm 0.0118$ & \textbf{0.6989} $\pm 0.0199$ \\
     &  & $0.6896 \pm 0.0670$ & $0.6676 \pm 0.0260$ & $0.6321 \pm 0.0874$ & \textbf{0.7016} $\pm 0.0483$ \\ \addlinespace
    
    \multirow{2}{*}{\begin{tabular}[c]{@{}c@{}}BRISC \\ (skull-stripped)\end{tabular}} & \checkmark & $0.6986 \pm 0.0393$ & $0.7091 \pm 0.0988$ & $0.7116 \pm 0.0479$ & \textbf{0.7248} $\pm 0.0163$ \\
     &  & $0.7222 \pm 0.0671$ & $0.6976 \pm 0.0109$ & $0.7005 \pm 0.0560$ & \textbf{0.7233} $\pm 0.0608$ \\
    \bottomrule
    
    \end{tabularx}
}
\end{table}

\subsubsection{2D Metadata-Conditioned Classification}
\begin{figure}[h]
\centering
\includegraphics[width=0.7\textwidth]{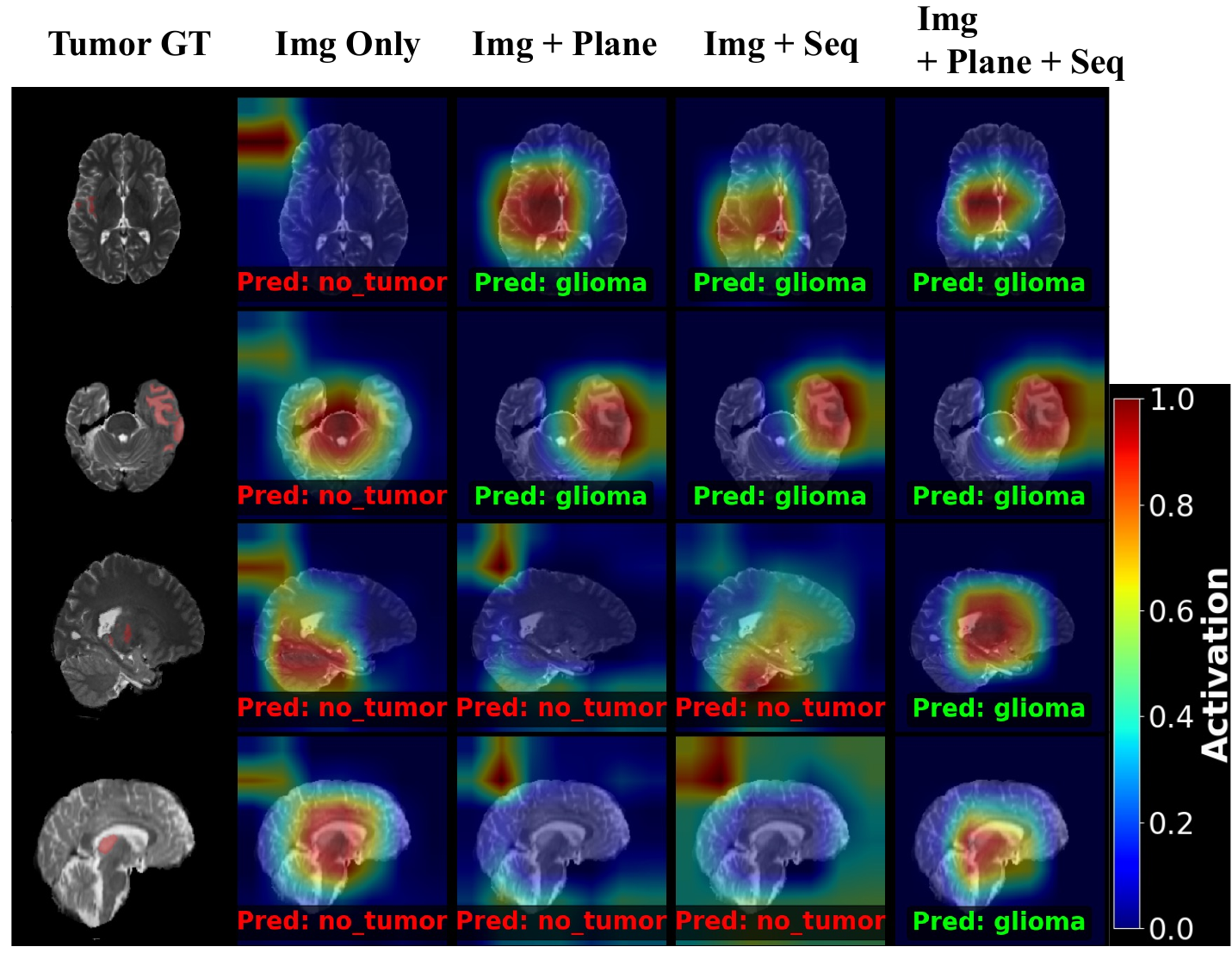}
\caption{Grad-CAM visualizations comparing the image-only baseline with our Meta-D architecture. The explicit integration of metadata successfully shifts the spatial attention pathways directly to the tumor margins, resolving contrast ambiguity.}
\label{fig:gradcam}
\end{figure}

Injecting spatial and sequence metadata consistently elevated performance above image-only baselines across all structural conditions, as shown in Table \ref{tab:2d_film}. The architecture consistently achieved top performance once both sequence and plane metadata were integrated simultaneously, particularly when N4 bias field correction~\cite{biascorrection} was applied to mitigate scanner variations. We performed permutation testing with randomized metadata during inference, inducing accuracy drops $\Delta$ Acc up to 10.28\%, confirming the active reliance of the model on explicit metadata rather than implicit visual cues. Concurrently, we observe that the mean magnitude of the scaling parameter $|\gamma|$ demonstrates the network actively scales intermediate feature representations by approximately 20\% based on sequence identity. Finally, Grad-CAM~\cite{gradcam} visualizations reveal that explicit routing shifts spatial attention directly to tumor margins (Fig.~\ref{fig:gradcam}).

\subsubsection{3D Segmentation under Missing Modalities}
Evaluated across all 15 scenarios on BraTS 2018~\cite{bratsref1,bratsref2,bratsref3} in Table \ref{tab:3d_brats}, Meta-D ($T_{max}$) universally outperformed the MMFormer~\cite{mmformer} baseline across every possible missing-modality combination. Improvements were highly pronounced under extreme structural degradation; for instance, relying solely on the T1 sequence yielded a substantial 5.12\% absolute increase in tumor segmentation Dice scores. By forcing attention weights of absent sequences to strictly zero through the metadata dictionary, the $T_{max}$ block mathematically prevents the extraction of noise from zero-padded regions and mitigates performance collapse.

\begin{table}[h!]
\centering
\caption{Comparison of average tumor segmentation Dice scores in percentages for all 15 missing scanning modality scenarios on BraTS 2018. The bullet symbol denotes an available modality and the open circle denotes a missing modality for FLAIR, T1c, T1, and T2. Best results per scenario are bolded.}
\label{tab:3d_brats}
\begin{adjustbox}{width=0.7\columnwidth} 
    \begin{tabular}{>{\centering\arraybackslash}p{1.3cm} >{\centering\arraybackslash}p{1.3cm} >{\centering\arraybackslash}p{1.3cm} >{\centering\arraybackslash}p{1.3cm} | c | c}
    \toprule
    \multicolumn{4}{c|}{\multirow{2}{*}{\textbf{Modalities}}} & \multirow{2}{*}{\textbf{3D Image only}} & \textbf{Meta-D (Ours)} \\
    \cmidrule(lr){6-6}
    \textbf{FLAIR} & \textbf{T1c} & \textbf{T1} & \textbf{T2} & & \textbf{Img + Seq} \\
    \midrule
    $\bullet$ & $\circ$ & $\circ$ & $\circ$ & 82.80 & \textbf{83.38} \\
    $\circ$ & $\bullet$ & $\circ$ & $\circ$ & 71.85 & \textbf{73.98} \\
    $\circ$ & $\circ$ & $\bullet$ & $\circ$ & 68.95 & \textbf{74.07} \\
    $\circ$ & $\circ$ & $\circ$ & $\bullet$ & 81.34 & \textbf{83.78} \\
    $\bullet$ & $\bullet$ & $\circ$ & $\circ$ & 85.74 & \textbf{86.27} \\
    $\bullet$ & $\circ$ & $\bullet$ & $\circ$ & 85.38 & \textbf{85.76} \\
    $\bullet$ & $\circ$ & $\circ$ & $\bullet$ & 86.56 & \textbf{87.00} \\
    $\circ$ & $\bullet$ & $\bullet$ & $\circ$ & 76.00 & \textbf{77.89} \\
    $\circ$ & $\bullet$ & $\circ$ & $\bullet$ & 85.04 & \textbf{85.74} \\
    $\circ$ & $\circ$ & $\bullet$ & $\bullet$ & 84.52 & \textbf{85.66} \\
    $\bullet$ & $\bullet$ & $\bullet$ & $\circ$ & 86.35 & \textbf{86.93} \\
    $\bullet$ & $\bullet$ & $\circ$ & $\bullet$ & 87.46 & \textbf{87.88} \\
    $\bullet$ & $\circ$ & $\bullet$ & $\bullet$ & 87.26 & \textbf{87.72} \\
    $\circ$ & $\bullet$ & $\bullet$ & $\bullet$ & 85.82 & \textbf{86.28} \\
    $\bullet$ & $\bullet$ & $\bullet$ & $\bullet$ & 87.72 & \textbf{88.24} \\
    \midrule
    \multicolumn{4}{c|}{\textbf{Average}} & 82.85 & \textbf{84.04} \\
    \bottomrule
    \end{tabular}
\end{adjustbox}
\end{table}

\subsubsection{Architectural Complexity}
Replacing standard spatial self-attention with our novel metadata-driven cross-attention explicitly routes spatial queries to a fixed metadata dictionary rather than redundant spatial locations (See Fig.~\ref{tab:efficiency}). This reduced the parameter count of the isolated bottleneck by 40.0\% and its computational burden by 50.0\%. Consequently, the end-to-end Meta-D achieved a 24.1\% reduction in total parameters and a 4.2\% reduction in GFLOPS as shown in Table \ref{tab:efficiency}. This demonstrates that explicitly routing spatial attention minimizes spatial parameterization overhead while maintaining segmentation robustness.

\begin{table}[t]
\centering
\caption{Computational complexity evaluated on a real $64 \times 64 \times 64$ BraTS MRI crop. We compare the end-to-end architecture and isolated attention bottleneck to demonstrate reductions from transitioning from spatial self-attention with complexity $O(N^2)$ to explicit metadata cross-attention with complexity $O(N)$.}
\label{tab:efficiency}
\begin{adjustbox}{width=0.8\textwidth}
\begin{tabular}{l  cc | cc}
\toprule
\multirow{2}{*}{\textbf{Model}} & \multicolumn{2}{c|}{\textbf{End-to-End Model}} & \multicolumn{2}{c}{\textbf{Isolated Bottleneck}} \\
\cmidrule(lr){2-3} \cmidrule(lr){4-5}
& \textbf{Params (M)} $\downarrow$ & \textbf{GFLOPS} $\downarrow$ & \textbf{Params (M)} $\downarrow$ & \textbf{GFLOPS} $\downarrow$ \\
\midrule
Baseline~\cite{mmformer} & 34.75 & 15.48 & 5.25 & 0.34 \\
Ours & \textbf{26.36} & \textbf{14.83} & \textbf{3.15} & \textbf{0.17} \\
\bottomrule
\end{tabular}
\end{adjustbox}
\end{table}

\section{Conclusion}

We have introduced Meta-D, an architecture for brain tumor analysis that leverages categorical scanning metadata to guide feature extraction. By replacing implicit feature inference with explicit metadata conditioning, Meta-D enables tumor detection and missing-modality tumor segmentation with measurable computational efficiency. Our results demonstrate that metadata-aware cross-attention prevents missing structural sequences from corrupting the latent feature space, yielding absolute performance gains of up to 5.12\% while reducing the total model parameters by 24.1\%. In the future, we plan to investigate integrating broader clinical variables and longitudinal scanner records to stabilize multi-parametric analysis. To encourage community adoption, we release Meta-D and our results as free and open research.

%
%
%
%
\bibliographystyle{splncs04}
\bibliography{references.bib}
\end{document}